\documentclass{article}

\usepackage[preprint]{airov24}

\usepackage[utf8]{inputenc} 
\usepackage[T1]{fontenc}    
\usepackage{hyperref}       
\usepackage{url}            
\usepackage{booktabs}       
\usepackage{amsfonts}       
\usepackage{nicefrac}       
\usepackage{microtype}      
\usepackage{xcolor}         

\usepackage{amsmath}
\usepackage[linesnumbered,ruled,algo2e]{algorithm2e}
\usepackage{leftidx}
\usepackage{graphicx}
\usepackage{tikz}
\usetikzlibrary{calc}
\usepackage{subcaption}
\usepackage{pgfplots}

\title{Workspace Registration and Collision Detection for Industrial Robotics Applications}

\author{%
Klaus Zauner\\
Institute of Robotics\\
Johannes Kepler University Linz\\
4040 Linz, Austria \\
\texttt{klaus.zauner@jku.at} \\
\And
Josef El Dib\\
Institute of Robotics\\
Johannes Kepler University Linz\\
\And
Hubert Gattringer\\
Institute of Robotics\\
Johannes Kepler University Linz\\
\texttt{hubert.gattringer@jku.at} \\
\AND
Andreas Müller\\
Institute of Robotics\\
Johannes Kepler University Linz\\
\texttt{a.mueller@jku.at} \\
}

\begin{document}

\maketitle

\begin{abstract}
Motion planning for robotic manipulators relies on precise knowledge of the environment in order to be able to define restricted areas and to take collision objects into account. To capture the workspace, point clouds of the environment are acquired using various sensors. The collision objects are identified by region growing segmentation and VCCS algorithm. Subsequently the point clusters are approximated. The aim of the present paper is to compare different sensors, to illustrate the process from detection to the finished collision environment and to detect collisions between the robot and this environment.
\end{abstract}

\section{Intoduction}\label{sec:intro}
The operation of robots in complex production environments requires precise knowledge of the latter. In addition to the definition of working areas and restricted areas, collision-free path and motion planning also plays an important role. To ensure smooth production processes, it is important to be able to automatically detect and take into account possible changes to the collision environment.\\

Various principles for capturing three-dimensional environmental information are available in the form of a wide range of sensors. The resulting point clouds require subsequent processing steps to create an environment representation by simple geometric objects. Many of the methods and algorithms used are freely available in the Point Cloud Library (PCL) \cite{Rusu_ICRA2011_PCL}. Time-optimal path planning requires the consideration of possible collisions in addition to the usual constraints for the optimization problem. A lot of research has been dedicated over the last few decades to theories and algorithms for detection collisions between objects \cite{ConvOptBoyd,Gilbert1988,GJK_Montanari,tracy2023differentiable}. Another crucial aspect is the incorporation of the resulting constraints into optimization problems for collision free motion planning. Several approaches dedicated to this problem have been published, e.\,g. \cite{Johnson1987,mueller04,chomp}.\\

This paper aims in comparing different sensors and illustrating the process from detection to the finished collision environment (section \ref{sec:detectionAndProcessing}). Moreover different approaches to object approximation (section \ref{sec:boundingBoxGeneration}) and collision detection shall be investigated in terms of usability of the actually available free space and the resulting number of constraints which must be taken into account for motion planning (section \ref{sec:collisionDetection}). An example illustrates the relevant relationships.

\section{Environment Description}\label{sec:environmentDescription}
A robotic manipulator can be described as compositions of rigid bodies $\mathcal{B}$. For each body $\mathcal{B}_i$ shall exist a body fixed frame $\mathcal{F}_i$ in which its spatial extent is described. With the robots geometry parameters $\mathbf{p}_{\text{geo}}$ and a vector $\mathbf{q}\in\mathbb{R}^n$ of $n$ joint coordinates, the position vectors $\leftidx{_I}{\mathbf{r}}{_{\mathcal{B}_i}}(\mathbf{q},\mathbf{p}_{\text{geo}})$, resolved in the inertial frame $\mathcal{F}_I$, can be calculated. The vector $\leftidx{_I}{\mathbf{r}}{_E}(\mathbf{q},\mathbf{p}_{\text{geo}})$ to the end-effector is of particular importance in robotics - it is also referred to as \textit{forward kinematics}.\\

The environment can be approximated in a similar way by cuboids, also with body fixed frame and location vector. However, these must first be defined. A viable approach of mapping a robot's environment is by a set of points $\mathbf{p}_i\in\mathbb{R}^3,\,i=1\,\dots\,N$, described in the reference frame $\mathcal{F}_I$, also referred to as point cloud (PC) and denoted as $\mathcal{P}^0$. Extracting the approximating cuboids out of this PC is the goal of the first part of this paper.\\

For the following considerations some properties of point clouds are required, which are introduced collectively here. Let $\mathbf{p}_i\in\mathbb{R}^3$ denote a particular point of the PC. The set of $k$ points next to it shall be referred to as k-neighborhood. The average distance between the point $\mathbf{p}_i$ and its neighboring points is calculated as
\begin{equation}
\label{eq:averageDistance}
d_i=\frac{1}{k}\sum\limits_{j=1}^k||\mathbf{p}_j-\mathbf{p}_i||_2.
\end{equation}
The second recurring property, that will be needed, is the covariance matrix
\begin{equation}
\label{eq:covMatrix}
\mathbf{C}=\frac{1}{k}\sum\limits_{j=1}^k\left(\mathbf{p}_j-\bar{\mathbf{p}}\right)^T\left(\mathbf{p}_j-\bar{\mathbf{p}}\right)\in\mathbb{R}^{3\times3},
\end{equation}
where $\bar{\mathbf{p}}$ denotes the mean value of the k-neighborhood. The eigenvalues $\lambda_l,\,l=1,2,3$ of the covariance matrix with the corresponding eigenvectors $\mathbf{v}_l,\,l=1,2,3$ indicate the orientation of the associated PC.

\section{Detection and Feature Extraxtion}\label{sec:detectionAndProcessing}
\subsection{Used 3D-Sensors}
In this paper, the \textit{time of flight} (TOF, e.\,g. Microsoft Kinect and Omron OS32C Lidar) method and the \textit{active stereoscopy} (e.\,g.Intel RealSense D435)  are used. Three different sensors are mounted at the endeffector of an industrial linear robot. These are then moved in the upper end position of the vertical axis, centered over the working area. During the movement, images are taken at equidistant intervals and then merged to form a PC covering the whole workspace.

\begin{figure}
\begin{subfigure}{0.45\linewidth}
\includegraphics[width=0.9\textwidth]{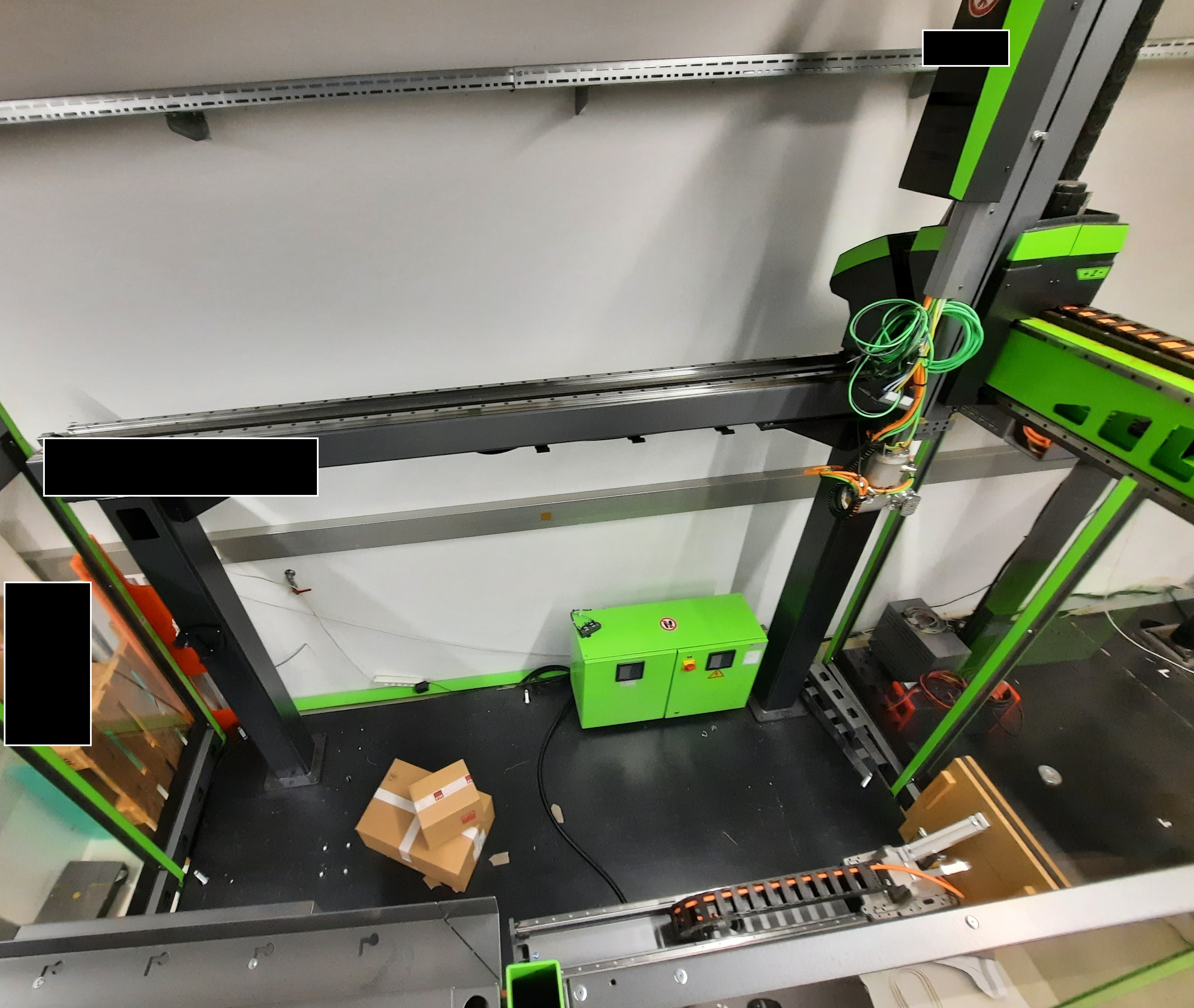}
\caption{Workspace of an industrial linear robot with several obstacles} \label{fig:workspace}
\end{subfigure}%
\hspace*{\fill}   
\begin{subfigure}{0.45\linewidth}
\includegraphics[width=0.9\textwidth]{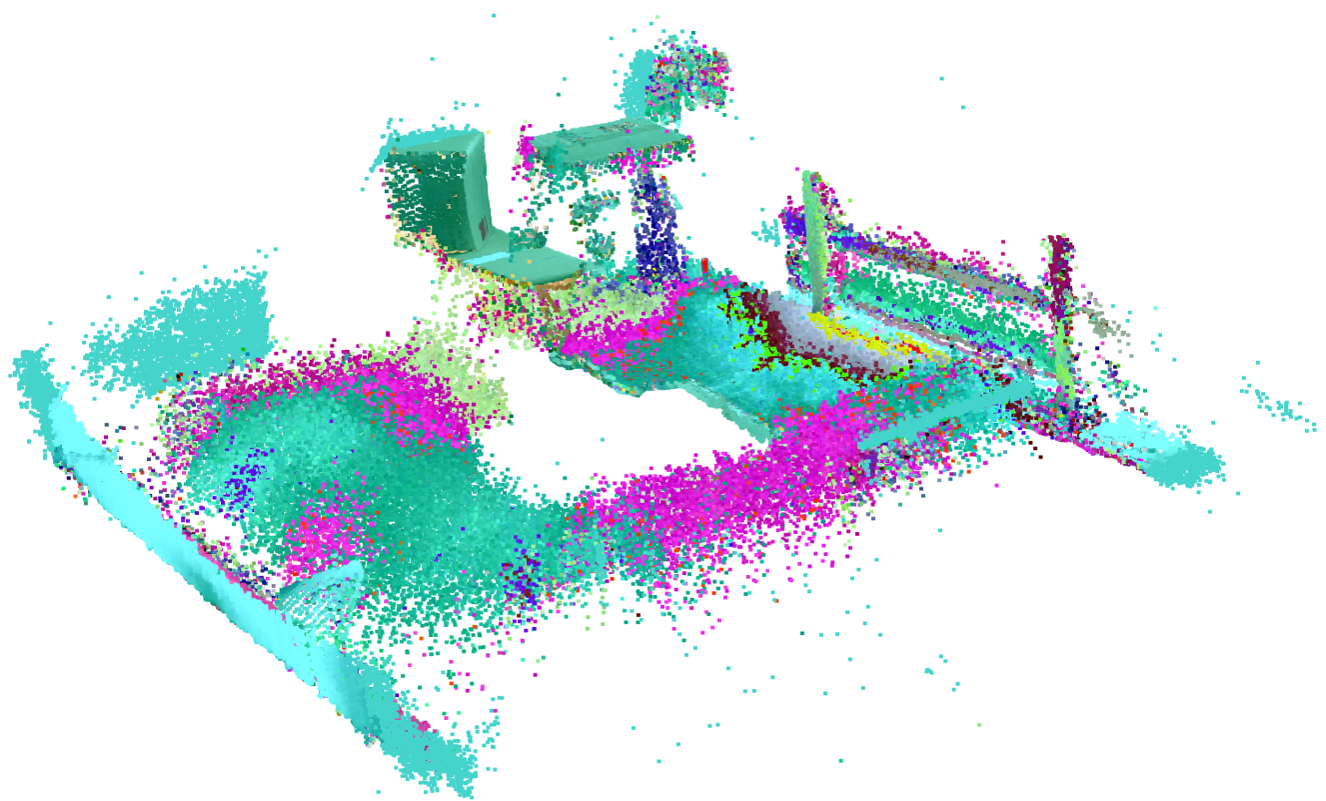}
\caption{PC representation of the workspace, sensor: Kinect V2} \label{fig:KinectV2_PC}
\end{subfigure}%
\caption{Comparison of sensors} \label{fig:comp1}
\end{figure}

\begin{figure}
\begin{subfigure}{0.45\linewidth}
\includegraphics[width=0.9\textwidth]{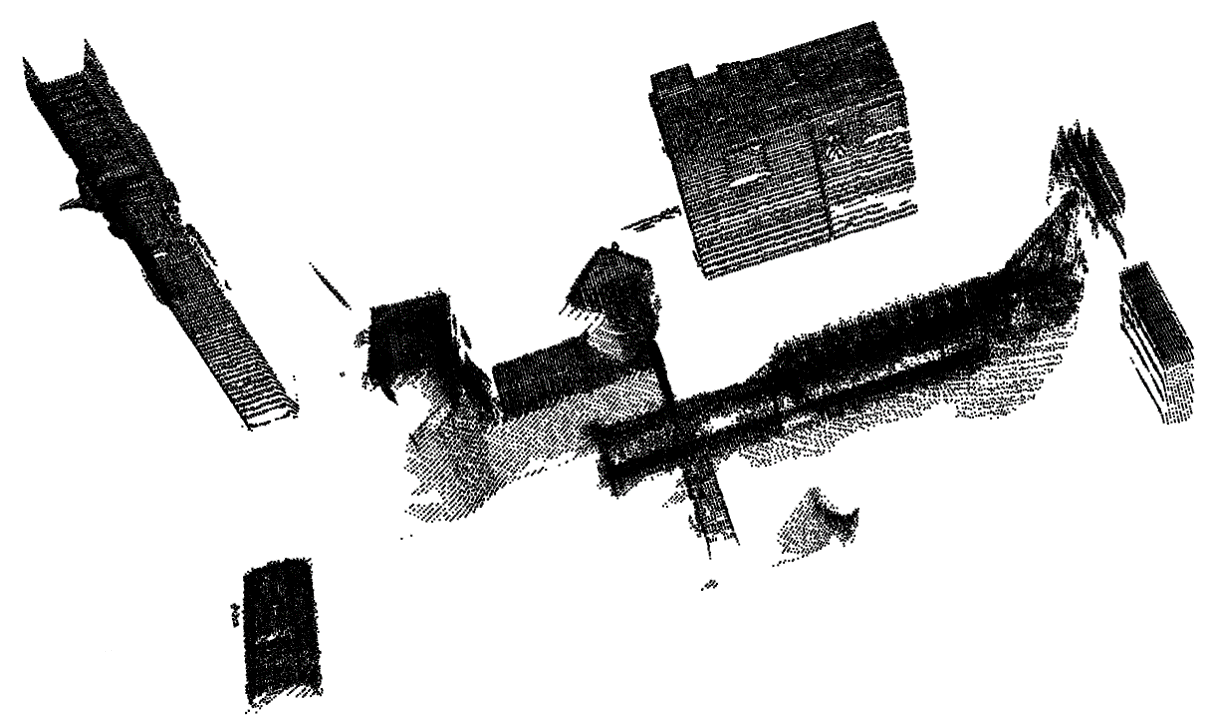}
\caption{PC representation of the workspace, sensor: Omron OS32C Lidar} \label{fig:Lidar_PC}
\end{subfigure}%
\hspace*{\fill}   
\begin{subfigure}{0.45\linewidth}
\includegraphics[width=0.9\textwidth]{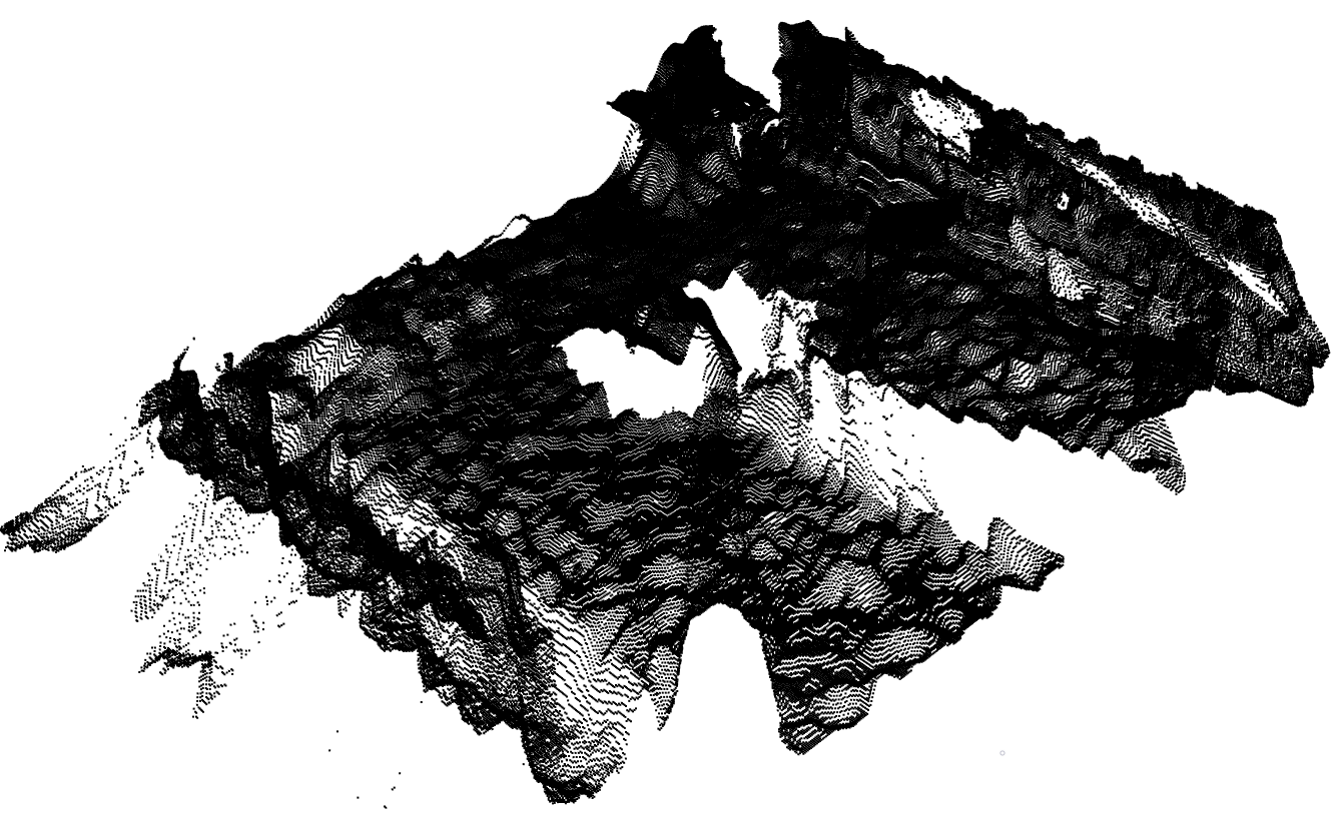}
\caption{PC representation of the workspace, sensor: Intel RealSense D435} \label{fig:RealSense_PC}
\end{subfigure}%
\caption{Comparison of sensors} \label{fig:comp2}
\end{figure}

Figures \ref{fig:KinectV2_PC} and \ref{fig:Lidar_PC} clearly show that although both Kinect and Lidar deliver good results, the more expensive Lidar has sharper edges and significantly less noise. The RealSense results in Fig. \ref{fig:RealSense_PC} show that no sharp transitions can be mapped and surfaces have a strong waviness, which can lead to a considerable loss of working space.

\subsection{Preprocessing}
Depending on the sensor resolution and the number of merged images, the PC can have a large number of points. Operating such large datasets makes it adviseable to use a kd tree \cite{kdTree}.\\

The PC will also cover areas outside the robots reachable space, which is determined as the image of $\leftidx{_I}{\mathbf{r}}{_E}(\mathbf{q},\mathbf{p}_{\text{geo}})$ under $\mathbf{q}$. Therefore the points outside the reachable space can be discared. The remaining points do not necessarily correspond to actual obstacles as they might contain noise - this depends on the used sensor. Removing as much of that noise as possible further reduces the amount of points to be processed and more importantly results in a larger useable workspace. Solid objects yield a certain local density of detected points. Assuming that the average distance (\ref{eq:averageDistance}) is gaussian distributed with mean $\bar{d}$ and standard deviation $\sigma$, every point $\mathbf{p}_i$ that lies outside $\bar{d}\pm u\sigma$, with the denoising coefficient $u$, is discarded as a statistical outlier. This gives the method its name - \textit{statistical outlier removal} \cite{SOT}. The pruned and filtered dataset is denoted as $\mathcal{P}$.
\subsection{Segmentation and Clustering}
For segmenting the PC a few more properties are required. A so-called cluster is a collection of points $\mathbf{C}_1$ around a seed point $\mathbf{p}_1\in\mathcal{P}$, which has a minimum distance $d_{th}$ to a set of points $\mathbf{C}_2$ around another seed point $\mathbf{p}_2\in\mathcal{P}$. Mathematically this states as
\begin{equation}
\label{eq:Clusters}
\text{min}\,||\mathbf{p}_i-\mathbf{p}_j||\geq d_{th},\,\mathbf{p}_i\in\mathbf{C}_1,\,\mathbf{p}_j\in\mathbf{C}_2.
\end{equation}
The points represent a surface. Therefore there has to be a direction in which the variance is significantly lower that in the other directions. Recalling the covariance matrix (\ref{eq:covMatrix}), this direction is given by the eigenvector with the lowest eigenvalue, denoted by $\lambda_1$, which is therefore an estimate for the surface normal vector. Moreover the surface curvature
\begin{equation}
\label{eq:Curvature}
\kappa=\frac{\lambda_1}{\lambda_1+\lambda_2+\lambda_3}
\end{equation}
can be calculated with the eigenvalues $\lambda_l,\,l=1,2,3$ of (\ref{eq:covMatrix}).
If $\kappa\leq\kappa_{th}$ holds for a specified threshold, the respective points can be added to a cluster as they form a smooth surface. Adjacent clusters $\mathbf{C}_1$ and $\mathbf{C}_2$ can be merged if their surface normal vectors $\mathbf{n}_1$ and $\mathbf{n}_2$ fullfill the condition
\begin{equation}
\label{eq:CurvatureMatching}
\arccos{\langle\mathbf{n}_1,\mathbf{n}_2\rangle}\leq\alpha_{th}.
\end{equation}
with the threshold $\alpha_{th}$ for the angle between $\mathbf{n}_1$ and $\mathbf{n}_2$.
This condition states that individual clusters belong to a surface as long as the curvature does not exceed a certain value, i.e. there are no corners or edges. The merging of clusters according to Alg. \ref{alg:clustering} is referred to as \textit{region growing segmentation} \cite{MERZOUGUI20191046,POUX2022104250}.

\begin{algorithm2e}[h]
	\SetKwInOut{Input}{Input}
	\SetKwInOut{Output}{Output}
	
	\underline{function clustering} $(\mathcal{P})$
	
	\Input{preprocessed point cloud $\mathcal{P}$}
	\Output{set of clusters}
	insert $\forall\mathbf{p}_i\in\mathcal{P}$ in a kd-tree\\
	set up an empty list of clusters\\
	set a queue $\mathcal{Q}$ of points to be checked\\
	\For{$\mathbf{p}_i\in\mathcal{P}$}{
		insert $\mathbf{p}_i$ in $\mathcal{Q}$\\
		check neighboring points $\mathbf{p}_i^k$ if they meet conditions \ref{eq:Clusters}, \ref{eq:Curvature} and \ref{eq:CurvatureMatching}\\
		\If{$\mathbf{p}_i^k$ not yet processed}
		    {insert $\mathbf{p}_i^k$ in $\mathcal{Q}$}
	}
	insert $\mathcal{Q}$ as cluster in the list of clusters\\
	empty $\mathcal{Q}$
	\caption{Region Growing Segmentation}\label{alg:clustering}
\end{algorithm2e}

\begin{figure}
\begin{subfigure}{0.5\linewidth}
\includegraphics[height = 3.5cm]{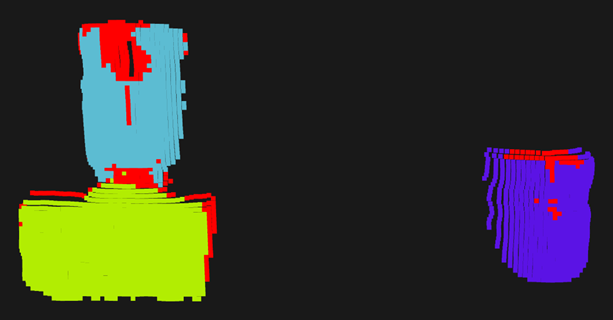}
\caption{Region growing segmentation}\label{fig:exRegionGrowingSegmentation}
\end{subfigure}%
\hspace*{\fill}   
\begin{subfigure}{0.5\linewidth}
\includegraphics[height = 3.5cm]{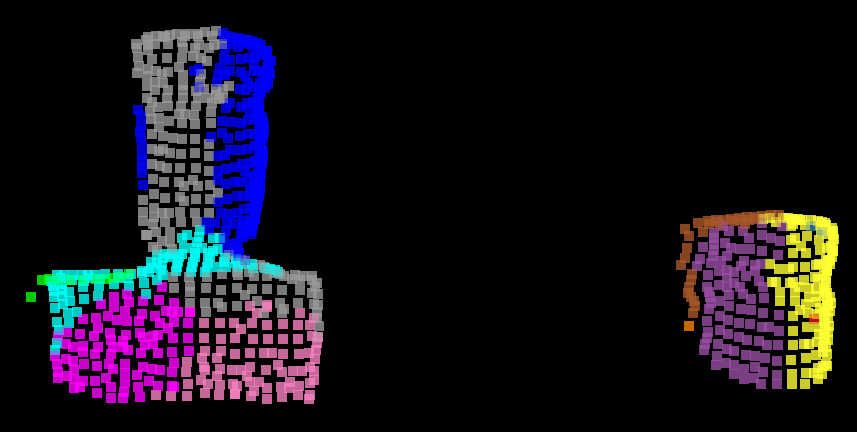}
\caption{VCCS} \label{fig:exVCCS}
\end{subfigure}%
\caption{Clustered example scenery}\label{fig:ClusteredExample}
\end{figure}

The proposed algorithm might result in large clusters, see Fig. \ref{fig:exRegionGrowingSegmentation} , that need to be split. This is of great importance as the subsequent approximation of very large clusters can lead to actually free workspace being regarded as part of an obstacle, see Fig. \ref{fig:BoundingBoxes}. One approach is using the \textit{supervoxel algorithm}, also known as the Voxel Cloud Connectivity Segmentation (VCCS) \cite{VCCS}. This results in smaller clusters, also referred to as supervoxels, see Fig. \ref{fig:ClusteredExample}.

\section{Bounding Box Generation}\label{sec:boundingBoxGeneration}
\begin{figure}[ht]
\begin{center}
\begin{tikzpicture}[]

\coordinate (origin1) at (0,0);
\coordinate (xPlus1) at (4,0);
\coordinate (yPlus1) at (0,4);
\draw[-latex, line width = 0.25mm] ($(origin1)+(-0.5,0)$) -- (xPlus1);
\draw[-latex, line width = 0.25mm] ($(origin1)+(0,-0.5)$) -- (yPlus1);
\node [] at ($(xPlus1)+(0,-0.25)$) {$\leftidx{_I}{x}{}$};
\node [] at ($(yPlus1)+(-0.25,0)$) {$\leftidx{_I}{y}{}$};

\coordinate (s1) at (0.2,2);
\coordinate (s2) at (0.7,1.4);
\coordinate (s3) at (1,1.3);
\coordinate (s4) at (2,1.5);
\coordinate (s5) at (2,0.8);
\coordinate (s6) at (3,0.5);
\coordinate (s7) at (2,0.4);
\coordinate (s8) at (1.2,0.8);
\coordinate (s9) at (0.8,2.5);
\coordinate (s10) at (1.5,2);
\coordinate (s11) at (1.3,1.1);
\coordinate (s12) at (2.5,1.3);
\coordinate (s13) at (3,1.3);
\coordinate (s14) at (2.3,1.5);
\coordinate (s15) at (2.2,2);
\coordinate (s16) at (3,1);

\filldraw [] (s1) circle (0.5mm);
\filldraw [] (s2) circle (0.5mm);
\filldraw [] (s3) circle (0.5mm);
\filldraw [] (s4) circle (0.5mm);
\filldraw [] (s5) circle (0.5mm);
\filldraw [] (s6) circle (0.5mm);
\filldraw [] (s7) circle (0.5mm);
\filldraw [] (s8) circle (0.5mm);
\filldraw [] (s9) circle (0.5mm);
\filldraw [] (s10) circle (0.5mm);
\filldraw [] (s11) circle (0.5mm);
\filldraw [] (s12) circle (0.5mm);
\filldraw [] (s13) circle (0.5mm);
\filldraw [] (s14) circle (0.5mm);
\filldraw [] (s15) circle (0.5mm);
\filldraw [] (s16) circle (0.5mm);

\draw [line width = 0.5mm, blue] (0.2,0.4) rectangle (3,2.5);

\begin{scope} [rotate=45]
\draw [line width = 0.5mm, green] (1.4,-1.8) rectangle (3.05,1.3);
\end{scope}

\draw [line width = 0.5mm, blue] (3,3.6) -- node [midway, above] {\textcolor{black}{axes aligned}} (4,3.6);
\draw [line width = 0.5mm, green] (3,3) -- node [midway, above] {\textcolor{black}{object aligned}} (4,3);

\end{tikzpicture}
\caption{PC with axis aligned and object aligned bounding boxes}
\label{fig:BoundingBoxes}
\end{center}
\end{figure}
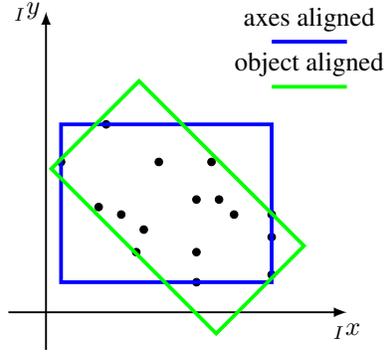

The clusters are approximated using cuboids in order to be able to use them efficiently for the definition of restricted areas or for collision detection. A viable way to do that is to place the boxes aligned to the reference frame $\mathcal{F}_I$. A disadvantage of this approach is that it might result in the the loss of actually usable workspace.  To better position the box according to the actual object it can be aligned with the principal axes of the cluster's covariance matrix (\ref{eq:covMatrix}), see Fig. \ref{fig:BoundingBoxes}.

\begin{figure}[ht]
\centering
\includegraphics[height = 5cm]{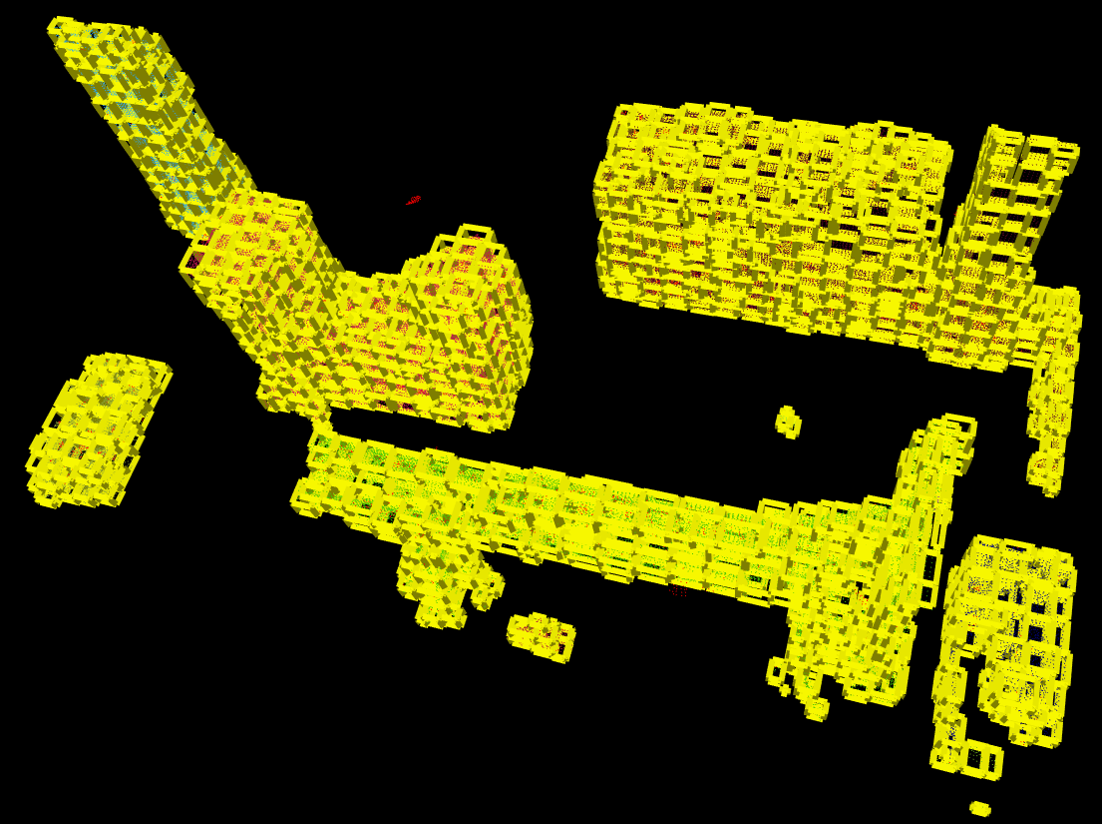}
\caption{Approximation of the robot environment Fig. \ref{fig:workspace}}
\label{fig:WorkspaceBoundingBoxes}
\end{figure}

Fig. \ref{fig:WorkspaceBoundingBoxes} shows the approximation of the robot environment Fig. \ref{fig:workspace} (captured with the Omron OS32C Lidar) with bounding boxes.

\section{Collision Detection}\label{sec:collisionDetection}
An essential part of motion planning for industrial robotics applications is the detection and prevention of collisions with the environment. This can be broken down to determining whether two bodies $\mathcal{B}_1$ and $\mathcal{B}_2$ have an intersection $\mathcal{B}_1\cap\mathcal{B}_2\neq\emptyset$ or not.

\subsection{Seperating Axis Theorem (SAT)}
Two three-dimensional objects $\mathcal{B}_1$ and $\mathcal{B}_2$ that do not collide can be spatially separated by a plane with normal vector $\mathbf{v}$. The normal vector is also called the separating axis. Therefore two objects do not collide if a vector can be found for which
\begin{equation}
\label{eq:seperating AxisTheorem}
\langle\mathbf{y},\mathbf{v}\rangle\geq c\,\text{and}\,\langle\mathbf{x},\mathbf{v}\rangle\leq c,
\end{equation}
for all $\mathbf{x}\in\mathcal{B}_1$, $\mathbf{y}\in\mathcal{B}_2$ and a real number $c$ hold \cite{ConvOptBoyd}. In words, this means that the projections of the bodies on the separating axis do not intersect. This theorem is only applicable for convex objects.\\

For two spherical objects, the separating axis can be specified directly as the connection between their centerpoints. The projection is then the radius of the sphere plotted from the center of the respective object. Let $O_1$ and $O_2$  be two objects approximated by bounding spheres of radius $r_1$ with center point $\mathbf{r}_1$ and $r_2$ with center point $\mathbf{r}_2$, respectively. Taking into account a safety margin $\delta$ the non-collision condition is
\begin{equation}
\label{eq:collDetectSpheres}
||\mathbf{r}_2-\mathbf{r}_1||_2\geq r_1+r_2+\delta.
\end{equation}
In terms of calculation this condition is of low complexity but the underlying method does have the disadvantage that a lot of usable working space is lost due to the spherical approximation.

\subsection{Minkowski-difference Based Collision Detection}
For the following, the objects are restricted to  be approximated by convex polyhedra. A convex polyhedron $\{\mathcal{V},\mathcal{E},\mathcal{F}\}$ with its vertices $\mathcal{V}$, edges $\mathcal{E}$ and faces $\mathcal{F}$ bounds a convex set in $\mathbb{R}^3$. The minimum distance of two convex polyhedra $\mathbf{A}$ and $\mathbf{B}$ is measured  between two witness points $\mathbf{a}\in\mathbf{A}$ and $\mathbf{b}\in\mathbf{B}$, which is not necessarily unique. Therefore the concepts of the \textit{physical space} and the \textit{configuration space} as introduced in \cite{GJK_Montanari} are used. In the configuration space, the spatial relation between two objects is given by their respective Minkowski difference.\\

Let $\mathbf{A}\subset\mathbf{V}$ and $\mathbf{B}\subset\mathbf{V}$ two subsets of a vector space $\mathbf{V}$. The Minkowski difference is constructed as
\begin{equation}
\mathbf{A}-\mathbf{B}:=\{\mathbf{a}-\mathbf{b}|\mathbf{a}\in\mathbf{A},\,\mathbf{b}\in\mathbf{B}\}.
\end{equation}
If $\mathbf{A}$ and $\mathbf{B}$ are bounded by convex polyhedra the Minkowski Difference is again bounded by a convex polyhedron.\\

\pgfplotsset{
compat=newest,
/pgfplots/myylabel absolute/.style={%
  /pgfplots/every axis y label/.style={at={(0,0.5)},xshift=#1,rotate=90},
  /pgfplots/every y tick scale label/.style={
    at={(0,1)},above right,inner sep=0pt,yshift=0.3em
   }
  }
}

\definecolor{mycolor}{rgb}{0.23935,0.30085,0.54084}%

\definecolor{c1}{rgb}{0,      0.4470,	0.7410}%
\definecolor{c2}{rgb}{0.8500, 0.3250,	0.0980}%
\definecolor{c3}{rgb}{0.9290, 0.6940,	0.1250}%
\definecolor{c4}{rgb}{0.4940, 0.1840,	0.5560}%
\definecolor{c5}{rgb}{0.4660, 0.6740,	0.1880}%
\definecolor{c6}{rgb}{0.6350, 0.0780,	0.1840}%

\def\myWidth{8cm}
\def\myHeight{1.3cm}

\begin{figure}[ht]
\centering
\subfloat[Colliding objects]{
\begin{tikzpicture}
\begin{axis}[
	name = plot1,
	width= 5cm,
	height= 5cm,
	scale only axis,
          axis equal,
	xmin=-4,
	xmax=4,
	ymin=-4,
	ymax=4,
	xlabel style={align=center,font=\color{white!15!black}},
	xlabel={x},
	ylabel style={align=center,font=\color{white!15!black}},
	ylabel={y},
	myylabel absolute=-15pt,
	axis background/.style={fill=white},
	xmajorgrids,
	ymajorgrids,
	xlabel style={font=\footnotesize},
	ylabel style={align=center,font=\footnotesize},
	ticklabel style={font=\footnotesize}, set layers,
	legend style={
		at={(0.45,1.2)},
		anchor=north,
		font=\footnotesize,
	legend columns=3,
	legend transposed=false},
	]
\addplot+[color=c1, line width=1pt, mark=none] coordinates {(1,1) (1,4) (3,4) (3,1) (1,1)};
\addlegendentry{$\mathbf{A}$};
\addplot+[color=c2, line width=1pt, mark=none] coordinates {(0,2) (0,3) (2,2) (0,2)};
\addlegendentry{$\mathbf{B}$};
\addplot+[color=c3, line width=1pt, mark=none] coordinates {(-1,-1) (-1,2) (3,2) (3,-2) (1,-2) (-1,-1)};
\addlegendentry{Minkowski difference};
\addplot+ [
mark = square*,
only marks
] coordinates {
(-1,-1)
(-1,2)
(1,-2)
(1,-1)
(1,1)
(1,2)
(3,-2)
(3,-1)
(3,1)
(3,2)
};
\addplot+ [
mark = o,
only marks,
mark size=4pt
] coordinates {
(0,0)
};
\label{fig:objColliding}
\end{axis}
\end{tikzpicture}
}\hfil
\subfloat[Spatially seperated objects]{
\begin{tikzpicture}
\begin{axis}[
	name = plot2,
	width= 5cm,
	height= 5cm,
	scale only axis,
          axis equal,
	xmin=-4,
	xmax=4,
	ymin=-4,
	ymax=4,
	xlabel style={align=center,font=\color{white!15!black}},
	xlabel={x},
	ylabel style={align=center,font=\color{white!15!black}},
	ylabel={y},
	myylabel absolute=-15pt,
	axis background/.style={fill=white},
	xmajorgrids,
	ymajorgrids,
	xlabel style={font=\footnotesize},
	ylabel style={align=center,font=\footnotesize},
	ticklabel style={font=\footnotesize}, set layers,
	legend style={
		at={(0.45,1.2)},
		anchor=north,
		font=\footnotesize,
	legend columns=3,
	legend transposed=false},
	]
\addplot+[color=c1, line width=1pt, mark=none] coordinates {(1,1) (1,4) (3,4) (3,1) (1,1)};
\addlegendentry{$\mathbf{A}$};
\addplot+[color=c2, line width=1pt, mark=none] coordinates {(0.5,-1) (0,2) (0.5,3) (0.5,-1)};
\addlegendentry{$\mathbf{B}$};
\addplot+[color=c3, line width=1pt, mark=none] coordinates {(0.5,-2) (0.5,5) (2.5,5) (3,2) (3,-1) (2.5,-2) (0.5,-2)};
\addlegendentry{Minkowski difference};
\addplot+ [
mark = square*,
only marks
] coordinates {
(0.5,-2)
(0.5,1)
(0.5,-2)
(0.5,5)
(1,-1)
(1,2)
(2.5,-2)
(2.5,1)
(2.5,2)
(2.5,5)
(3,-1)
(3,2)
};
\addplot+ [
mark = o,
only marks,
mark size=4pt
] coordinates {
(0,0)
};
\label{fig:objNotColliding}
\end{axis}
\end{tikzpicture}
}
\caption{Illustration of the concept of collision detection based on the Minkowski difference}
\label{fig:collisionMinkowski}
\end{figure}
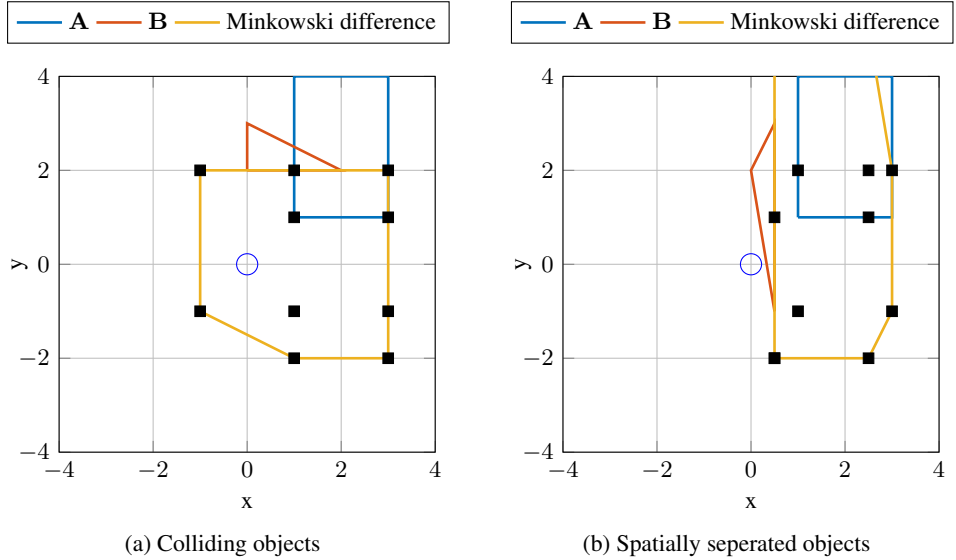

With two bodies represented by convex polyhedra the question arises whether they collide or not and what is the penetration depth or the minimum distance, respectively. The former can be answered by checking if there are points shared by both objects or equivalently if the origin lies within the Minkowski difference, see Fig. \ref{fig:collisionMinkowski}.\\

One widely used algorithm  is the Gilbert-Johnson-Keerthi (GJK) algorithm originally published in \cite{Gilbert1988} and repeatedly modified over the years, for example in as presented in \cite{GJK_Montanari}. Knowing whether two objects collide or not might be insuficient for some applications as \textit{collision-free motion planning} where a gradient and therefore a configuration dependent penetration depth is needed in order to move the path towards a feasible solution. The GJK algorithm itself does not provide with the penetration depth, but there exist algorithms to do that, for example the expanding polytope algorithm (EPA) \cite{feng2020minkowski}.

\subsection{Method comparison}\label{subsec:example}
A significant difference between the two methods is that the boxes (or other forms of convex polytropes) can be used directly processed by the GJK algorithm, while the application of the separating axis theorem in the simplified form (\ref{eq:collDetectSpheres}) requires the approximation of the boxes by spheres. The latter can lead to a considerable loss of free space.\\

\begin{figure}
\begin{subfigure}{0.45\linewidth}
\includegraphics[width=1\textwidth]{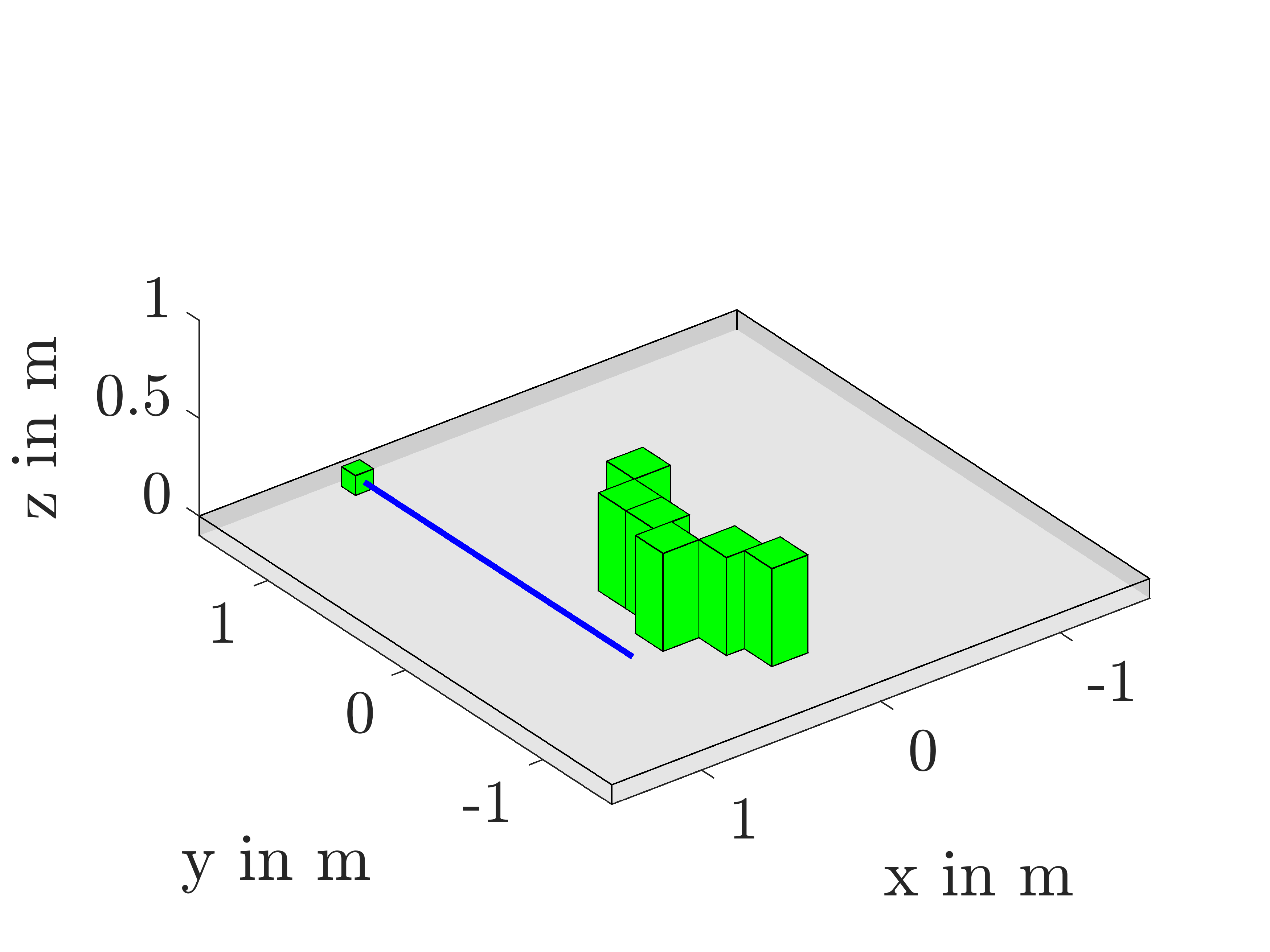}
\caption{Setup}\label{fig:exampleSetup}
\end{subfigure}%
\hspace*{\fill}   
\begin{subfigure}{0.45\linewidth}
\begin{tikzpicture}
\begin{axis}[
	name = plot2,
	width= 5cm,
	height= 3cm,
	scale only axis,
          axis equal,
	xmin=-1,
	xmax=1,
	ymin=0,
	ymax=1.5,
	xlabel style={align=center,font=\color{white!15!black}},
	xlabel={y in m},
	ylabel style={align=center,font=\color{white!15!black}},
	ylabel={min. dist. in m},
	myylabel absolute=-25pt,
	axis background/.style={fill=white},
	xmajorgrids,
	ymajorgrids,
	xlabel style={font=\footnotesize},
	ylabel style={align=center,font=\footnotesize},
	ticklabel style={font=\footnotesize}, set layers,
	legend style={
		at={(0.5,1.4)},
		anchor=north,
		font=\footnotesize,
	legend columns=2,
	legend transposed=false},
	]
\addplot+[solid, color=black, line cap=round, mark=none, line width=1pt, on layer=axis foreground, restrict x to domain=-1:1] table[x=x, y=distGJK, col sep=semicolon] {file1.csv};
	\addlegendentry{GJK}
\addplot+[solid, color=c1, line cap=round, mark=none, line width=1pt, on layer=axis foreground, restrict x to domain=-1:1] table[x=x, y=distSphere, col sep=semicolon] {file1.csv};
	\addlegendentry{SAT, $N=1$}
\addplot+[solid, color=c2, line cap=round, mark=none, line width=1pt, on layer=axis foreground, restrict x to domain=-1:1] table[x=x, y=distSphere, col sep=semicolon] {file2.csv};
	\addlegendentry{SAT, $N=2$}
\addplot+[solid, color=c3, line cap=round, mark=none, line width=1pt, on layer=axis foreground, restrict x to domain=-1:1] table[x=x, y=distSphere, col sep=semicolon] {file6.csv};
	\addlegendentry{SAT, $N=6$}
\end{axis}
\end{tikzpicture}
\caption{Minimum distance to an obstacle for GJK and SAT with $N=1,2,6$ spheres each}\label{fig:compareGJKandSAT}
\end{subfigure}%
\caption{Example}\label{fig:example}
\end{figure}

To illustrate this, a cubic object is moved past a row of objects, as depicted in Fig. \ref{fig:exampleSetup}, and the minimum distance is calculated - once with the GJK algorithm and once with condition (\ref{eq:collDetectSpheres}). For the latter, the approximation of each cuboid is carried out with $N=1$, $N=2$ and $N=6$ spheres each. Fig. \ref{fig:compareGJKandSAT} shows the results.

Possible applications for the collision-detection are online safety measurements for emergency stoping an ongoing movement in order to prevent damage to man and/or machine, as well as the task of collision free path planning. The latter relies on a good trade off between approximation accuracy and a reasonable number of constraints. For example the collision free movement from a point $\mathbf{r}_A$ to a point $\mathbf{r}_B$ on a \textit{B-Spline} parametrized path $\leftidx{_I}{\mathbf{r}}{_E}=\mathbf{z}_E(\mathbf{d},t)$ with the vector of control points $\mathbf{d}$ and the time $t\in[t_0,t_e]$ as path parameter, gives rise to the optimization problem
\begin{equation}
\begin{array}{rc}
 & \underset{\mathbf{d},t_e}{\text{min}}\,J(.)=\int\limits_{t_0}^{t_E}\left(\alpha+\beta\dddot{\mathbf{z}}_E^T\dddot{\mathbf{z}}_E\right)dt\\
\text{s.t.} & \underline{\mathbf{q}}\leq\mathbf{q}(\mathbf{d},t)\leq\bar{\mathbf{q}}\\
& \underline{\dot{\mathbf{q}}}\leq\dot{\mathbf{q}}(\mathbf{d},t)\leq\bar{\dot{\mathbf{q}}}\\
& \underline{\ddot{\mathbf{q}}}\leq\ddot{\mathbf{q}}(\mathbf{d},t)\leq\bar{\ddot{\mathbf{q}}}\\
& +\,\text{collision avoidance.}
\end{array}
\end{equation}
For every intermediate candidate $\mathbf{z}_E(\mathbf{d}_{\text{cand.}},t)$, it needs to be checked for discrete points on the path if every bounding box of the manipulator maintains a minimum distance to every bounding box of the environment. Let be $t_i,\,i=1,\,\dots,\,N$ the discrete points where the constraints for collision avoidance are evaluated. With the number of bounding boxes $n_{\text{rob}}$ for the robot and $n_{\text{env}}$ for the environment, one ends up with a total number
\begin{equation}
n_{\text{constr}}=n_{\text{rob}}n_{\text{env}}N
\end{equation}
of constraints to be evaluated during every iteration of the optimization process. Referring to Fig. \ref{fig:compareGJKandSAT} that would result in $n_{\text{rob}}n_{\text{env}}=36$ constraints for each discrete checkpoint to come even near the GJK algorithm where only $n_{\text{rob}}n_{\text{env}}=6$ constraints need to be evaluated per checkpoint.

\section{Summary and Outlook}
In the present paper we give an overview on the necessary steps from detecting an environment with different sensors and preprocessing the resulting point cloud to feature extraction and approximation with defined objects. Subsequently two methods for collision detection as well as their advantages and disadavantages are shown and compared in subsection \ref{subsec:example}. The main criteria are the number of bounding boxes and how the associated discretization affects the quality of the approximation and the remaining useable workspace.\\

As using the Minkowski-difference based collision detection allows to decrease both $n_{\text{rob}}$ and $n_{\text{env}}$, the number of constraints will be reduced to only a fraction of the necessary constraints using the spherical approximation. If that necessarily means lower computation time needs to be further investigated, as the condition (\ref{eq:collDetectSpheres}) is of lower complexity than the GJK algorithm.

\begin{ack}
This work has been supported by the "LCM – K2 Center for Symbiotic Mechatronics" within the framework of the Austrian COMET-K2 program.
\end{ack}

\bibliographystyle{plain}
\bibliography{airov_references_zauner}

\end{document}